\documentclass{styles/svproc}

\usepackage[space]{cite}

\usepackage{hyperref}
\usepackage{academicons}

\usepackage{hyperref} 

\usepackage{booktabs}
\usepackage{enumitem}
\usepackage{csquotes}
\usepackage{graphicx, multirow}
\usepackage[dvipsnames]{xcolor}
\usepackage{subfig}
\usepackage{comment}
\usepackage{wrapfig}
\usepackage{caption}
\definecolor{orcidlogocol}{HTML}{A6CE39}
\newcommand{\orcid}[1]{\href{https://orcid.org/#1}{\textcolor[HTML]{A6CE39}{\aiOrcid}}}

\usepackage{scalerel}
\usepackage{tikz}
\usetikzlibrary{svg.path}
\definecolor{orcidlogocol}{HTML}{A6CE39}
\tikzset{
    orcidlogo/.pic={
        \fill[orcidlogocol] svg{M256,128c0,70.7-57.3,128-128,128C57.3,256,0,198.7,0,128C0,57.3,57.3,0,128,0C198.7,0,256,57.3,256,128z};
        \fill[white] svg{M86.3,186.2H70.9V79.1h15.4v48.4V186.2z}
        svg{M108.9,79.1h41.6c39.6,0,57,28.3,57,53.6c0,27.5-21.5,53.6-56.8,53.6h-41.8V79.1z M124.3,172.4h24.5c34.9,0,42.9-26.5,42.9-39.7c0-21.5-13.7-39.7-43.7-39.7h-23.7V172.4z}
        svg{M88.7,56.8c0,5.5-4.5,10.1-10.1,10.1c-5.6,0-10.1-4.6-10.1-10.1c0-5.6,4.5-10.1,10.1-10.1C84.2,46.7,88.7,51.3,88.7,56.8z};
    }
}

\newcommand\orcidicon[1]{\href{https://orcid.org/#1}{\mbox{\scalerel*{
                \begin{tikzpicture}[yscale=-1,transform shape]
                \pic{orcidlogo};
                \end{tikzpicture}
            }{|}}}}

\usepackage{url}

\begin{document}
\mainmatter              
\title{A fully automated end-to-end process for fluorescence microscopy images of yeast cells: From segmentation to detection and classification}
\titlerunning{From Segmentation to Detection and Classification}  
%
\author{Asmaa Haja$^{\textsuperscript{\orcidicon{0000-0002-4116-2167}}}$ \and Lambert R.B. Schomaker$^{\textsuperscript{\orcidicon{0000-0003-2351-930X}}}$} 
\authorrunning{Asmaa Haja and Lambert R.B. Schomaker} 
%
\tocauthor{Asmaa Haja and Lambert R.B. Schomaker}
\institute{Bernoulli Institute, University of Groningen, The Netherlands\\
\email{ \{a.haja, l.r.b.schomaker\}@rug.nl }
}

\maketitle              

%
%
\begin{abstract}
In recent years, an enormous amount of fluorescence microscopy images were collected in high-throughput lab settings. 
Analyzing and extracting relevant information from all images in a short time is almost impossible. 
Detecting tiny individual cell compartments is one of many challenges faced by biologists. 
This paper aims at solving this problem by building an end-to-end process that employs methods from the deep learning field to automatically segment, detect and classify cell compartments of fluorescence microscopy images of yeast cells. 
With this intention we used Mask R-CNN to automatically segment and label a large amount of yeast cell data, and YOLOv4 to automatically detect and classify individual yeast cell compartments from these images. 
This fully automated end-to-end process is intended to be integrated into an interactive e-Science server in the PerICo\footnote{https://itn-perico.eu/home/} project, which can be used 
by biologists 
with minimized human effort in training and operation to complete their various classification tasks.
In addition, we evaluated the detection and classification performance of state-of-the-art YOLOv4 on data from the NOP1pr-GFP-SWAT yeast-cell data library. 
Experimental results show that by dividing original images into 4 quadrants YOLOv4 outputs good detection and classification results with an F1-score of 98\% in terms of accuracy and speed, which is optimally suited for the native resolution of the microscope and current GPU memory sizes.
Although the application domain is optical microscopy in yeast cells, the method is also applicable to multiple-cell images in medical applications.
\end{abstract}
%
%
%
%
\section{Introduction\label{sec:introduction}}
The existence of modern microscopy facilitates the generation of high-throughput data: It is now possible to produce very large collections of microscopic images of cell samples in a short time. The enormous amount of data opens the door for the biologists to study important and more complex aspects in their research field. However, one of many challenges they are recently facing is how to process such amount of data in a short time, extracting as much information as possible, as well as identifying biologically and clinically relevant diseases such as human diseases. Analyzing a huge volume of microscopy images by manually going though every image is a tedious task, can lead to fatigue and decision errors. Therefore, there is a desire to automatically process and analyse data in a high-throughput setting with minimized human effort in training and operation. Integrating techniques from the deep learning field of artificial intelligence seems to be a promising solution for this problem. Automatic detection and classification of details in microscopic images would dramatically speed up their research and contribute to their field of knowledge.\\
In this paper, our focus is on a specific problem in the field of biology, which is the automatic detection of individual-cell compartments in fluorescence microscopy images of yeast cells, notably organelle, as well as automatic specification of their type. 
In fact, the highlight of this paper is on the application of deep learning algorithms to biological data, i.e., images from optical cell microscopy. In this study we will not go into details to cover biological concepts.\\
Object detection and classification is one of the hottest topic in the deep learning field. Different approaches were developed for the detection, segmentation and classification of various cell types. In traditional approaches, each of these steps were implemented as separate algorithms. As an example, such approaches used morphology methods for detection \cite{colin2006detection, anoraganingrum1999cell}, whereas new approaches use machine learning and/or deep learning methods to realize these steps in a more realistic manner. Notably, Convolutional Neural Networks (CNN) are able to realize the same functionality using end-to-end training \cite{dong2015deep, pan2018cell}, as opposed to meticulous design of a processing pipeline with individual processing stages. In \cite{shitong2006new}, for instance, a morphological gray reconstruction based on a fuzzy cellular neural network is applied to detect white blood cells.
Xipeng et. al \cite{pan2018cell} proposed a novel multi-scale fully CNNs approach for regression of a density map to detect both nuclei of pathology and microscopy images. 
Xie et al. \cite{xie2018microscopy} developed two convolutional regression networks to detect and count cells.
Wang et al. \cite{wang2016subtype} in 2016 combined two CNN for simultaneously detecting and classifying cells.\\
Although there are many specialised methods that are capable of detecting different types of cells, to the best of the authors knowledge there exists no generic system for detecting all kinds of cell compartment in an accurate and easy way. In this paper, we present a fully automated end-to-end process for yeast-cell data that is capable of solving various segmentation, detection and classification tasks. For that, we use Mask-RCNN \cite{he2017mask} to automatically segment and label images from the input data, and YOLOv4 \cite{bochkovskiy2020YOLOv4} to automatically detect and classify individual yeast cell compartments from these images. This end-to-end process is currently intended to be integrated into an interactive e-Science server in the PerICo\footnote{https://itn-perico.eu/home/} project. We also evaluate the detection and classification performance of YOLOv4 on data from the NOP1pr-GFP-SWAT yeast-cell data library. We chose this particular algorithm because it is capable of detecting small objects, such as individual cell compartments, requiring a limited computation time. YOLO detects and classifies objects in only one stage, i.e., in one run. \\
The remainder of the paper is structured as follows: Section \ref{sec:Data} presents the used data. In section \ref{sec:End-to-end-process}, our end-to-end process is introduced. 
Section \ref{sec:Experimental_Design} provides an overview of the our experimental design. 
The results of our study on the chosen dataset is presented in section \ref{sec:Results_Discussion}, while the last section concludes the paper and indicates the future works. 
%
%
%
%
\section{Data\label{sec:Data}}
\begin{figure}[htb!]
\centering
\includegraphics[width=0.90\textwidth]{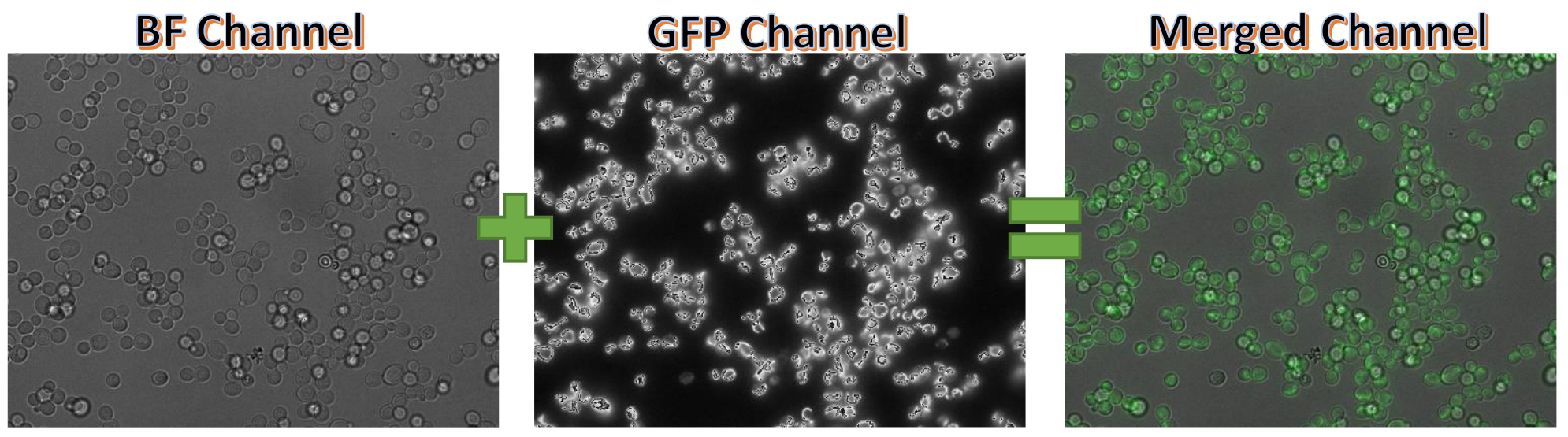}
    \caption{Merging BF and GFP channels of a randomly selected image [Plate15 J9]. 
    \label{FIG:3-channels}}
    \vspace{-17pt}
\end{figure}

To evaluate the end-to-end process we used publicly available fluorescence microscopy data from a library of yeast strains each expressing one protein under control of a constitutive promoter (NOP1) and fused to a Green Fluorescent Protein (GFP) at the N terminus (NOP1pr-GFP-SWAT library) \cite{weill2018genome}. This library contains annotated GFP and Bright Field (BF) yeast images of nearly 6000 strains each residing in a specific organelle in the cell. Overall, there were 18432 images from 16 well-plates, each consists of 1152 images for each channel with the dimension of 1344 x 1024 pixels. With respect to deep learning, each image is represented by a pre-defined class that describes the objects found in the image. Here, the classes are defined by the cell-compartments names. 
\begin{wraptable}{r}{0.6\textwidth}
\vspace{-22pt}
\caption{Number of unique images for cell compartments that have more than 300 images.}\label{tab:unique_organelle}
\centering
\begin{tabular}{lc}
  \toprule
  Name of cell compartments & \# Unique images \\
  \midrule
  ER (Endoplasmic Reticulum) & 376 \\
  Cytosol \& nucleus  & 401 \\
  Mitochondria & 461 \\
  Nucleus & 660 \\
  Cytosol & 1566 \\
  \bottomrule
\end{tabular}
\vspace{-24pt}
\end{wraptable}
Table \ref{tab:unique_organelle} lists the classes that have more than 300 unique images. In Figure \ref{FIG:3-channels}, merging a BF and GFP channels of a random sample image is shown. The BF is shown on the left side of this Figure, while GFP channel is shown in the middle side of the Figure. The results of merging both channels are shown in green and grey colors in the right side of Figure \ref{FIG:3-channels}. We chose this particular data because individual cells are too small to detect, it contains overlapped and close cells, and it consists of different cell sizes and shapes as can be seen in this Figure.
\vspace{-18pt}

%
%
%
%

\section{End-to-end Process\label{sec:End-to-end-process}}

\begin{figure*}[!htb]
\centering
\includegraphics[width=0.95\textwidth]{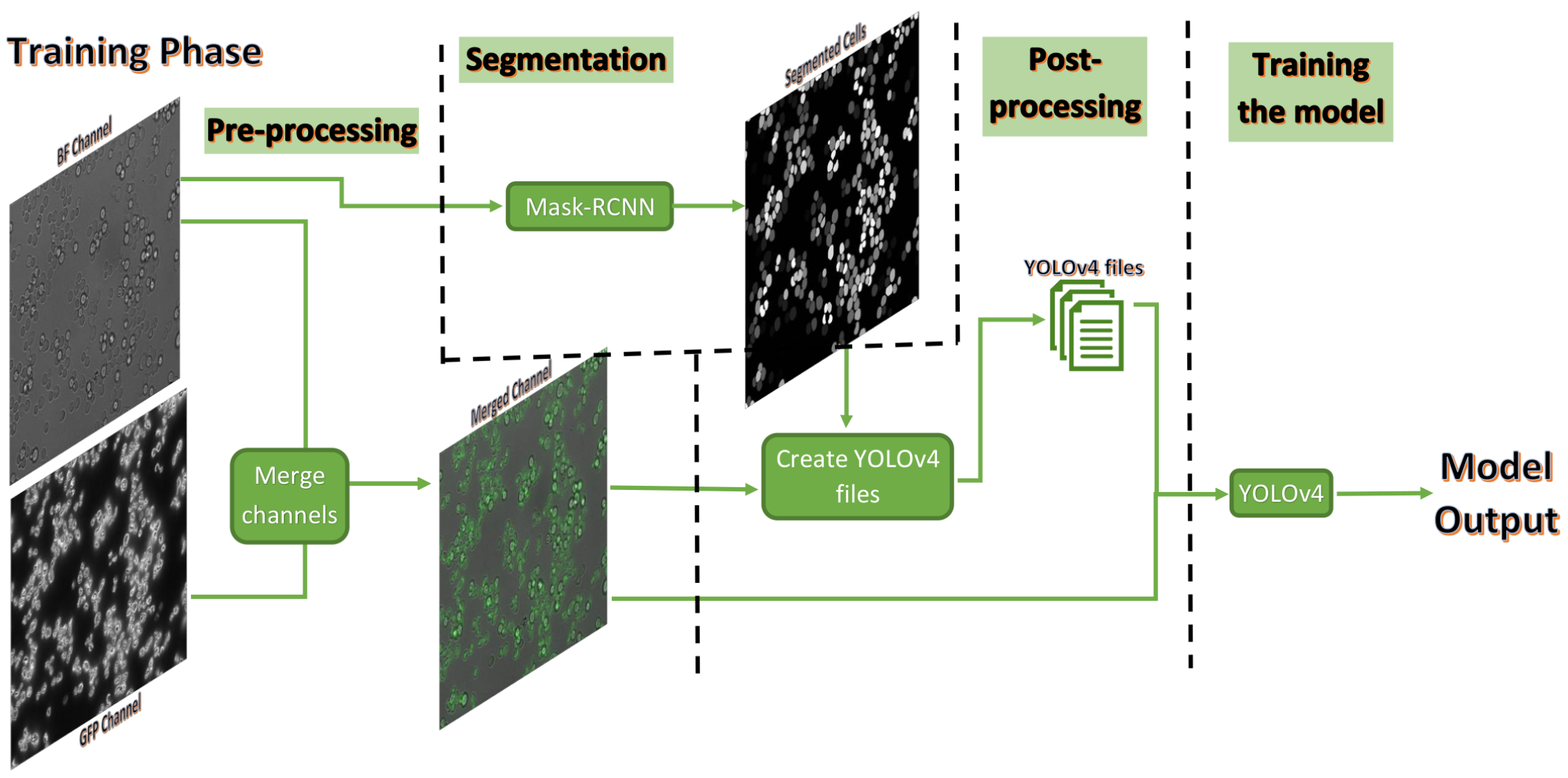}
    \caption{Pipeline of the training phase. 
    \label{FIG:TrainingPhase2}}
\end{figure*}

\begin{figure*}[!htb]
\centering
\includegraphics[width=0.95\textwidth]{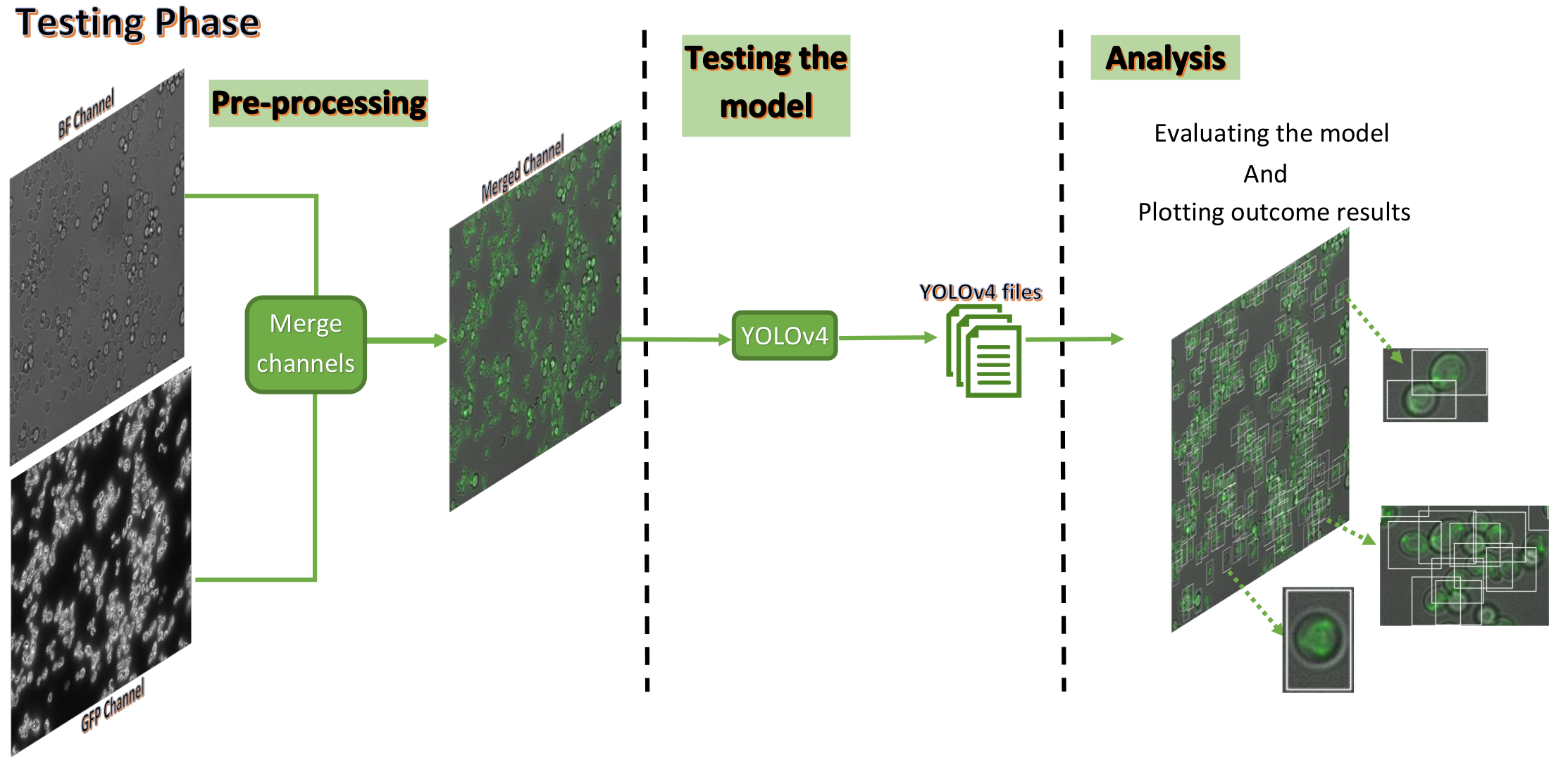}
    \caption{Pipeline of the testing phase including analysis. 
    \label{FIG:TestingPhase3}}
\end{figure*}

The main goal of this paper is to present the fully automated end-to-end segmentation, detection and classification process as well as to evaluate the individual-cell compartments detection performance of the state-of-the-art YOLOv4 model on fluorescence microscopy images. The traditional approach in the deep learning for carrying out any kind of task is to execute it in two phases: training- and testing phase. For that, the data will randomly be divided into two unmixed sets: train- and test sets, each contains the same amount of data for each class. In a training plus validation phase, we teach the model to detect individual-cell compartment by providing it with the different locations of almost all individual cells in the training images.
In order to test the trained model, we provide it with test images that the model did not see before. In the end, the model is evaluated based on specific metrics that decide how well the model has learned the specific tasks, the detection and classification tasks in this case. 
\\
Every image in our data is characterized by a label (cell-compartment name).
However, we are not only interested to know which type of cells can be found in the image but also the exact locations of each individual cell.
One of many state-of-the-art solutions to automatically segment individual objects is the mask-RCNN model proposed in \cite{he2017mask}. According to \cite{lu2019yeastspotter}, a pre-trained mask-RCNN model can be used to segmented yeast cells without fine-tuning. We used their implementation to segment individual cells in our dataset. It is to be noticed that the segmentation is completed in an unsupervised manner, done on the bright-field channel and not on the GFP channel. 
For the detection and classification of the individual-cell compartments we use the state-of-the-art YOLOv4 model developed by Bochkovskiy, Wang and Liao \cite{bochkovskiy2020YOLOv4}. The primary goal of their paper is to design a fast-operating object detector for production systems that is also optimized for parallel computations, and more importantly is that the training should be done on one single conventional GPU.
In comparison to other existing state-of-the-art models, YOLOv4 outperforms them in term of speed, accuracy and performance \cite{bochkovskiy2020YOLOv4}. 
Not only that but it seems a good candidate to use for detecting small objects seeing all modifications that were added to it, which are considered as a significant upgrade to its previous well-known versions. Therefore, we consider it as the best starting point for addressing the where and what question, i.e., detection and classification, in microscopic images. 
\\
Figure \ref{FIG:TrainingPhase2} shows the pipeline of the training phase. First, we use Mask-RCNN model to segment individual cells on each BF channel in the training set [segmentation step]. Simultaneously, we merge both BF and GFP channels for each image [pre-processing step]. With the purpose of training YOLOv4, we create specific YOLOv4 files from both the merged and the segmented images [post-processing step]. The last step in this phase would be to train YOLOv4 using both the created files and the merged images from the training set [training the model step]. In Figure \ref{FIG:TestingPhase3}, a pipeline of the testing phase is presented. Similar to the training phase, we first need to merge both BF and GFP channels from the unseen images in the test set [pre-processing step]. We use these images to test the trained YOLOv4 model [testing the model step]. As a result, YOLOv4 yields files for each test image, where the predicted location of each individual cell is computed. We use these files to evaluate the performance of YOLOv4 [analysis step]. 
The results described by the segmentation outcome of the training images, the trained model parameters, and the outcome of the detection and classification of the trained model on the test images are presented to the user.

\vspace{-17pt}
\vspace{-35pt}

\begin{table*}[htb!]
\caption{Experiment, image size, classes [ER, Mitochondria (M), Cytosol (C) and Nucleus(N)], number of images in train-, validation-, and test set.}\label{tab:images_total} \centering
\resizebox{\columnwidth}{!}{%
\begin{tabular}{lccccc}
  \toprule
  \# Experiment & Images size & Classes & \# Train images & \# Validation images  & \# Test images  \\
  \midrule
  $Exp_{1}$ & 1344 x 1024 & M & $\approx$ 990 & $\approx$ 110 & $\approx$ 270 \\
  $Exp_{2}$ & 1344 x 1024 & ER, M & $\approx$ 1620 & $\approx$ 180 & $\approx$ 450\\
  $Exp_{3}$ & 1344 x 1024 & ER, M, C, N & $\approx$ 2980 & $\approx$ 330 & $\approx$ 912 \\
  $Exp_{4}$ & 672 x 512 & M & $\approx$ 3960 & $\approx$ 440 & $\approx$ 1110\\
  $Exp_{5}$ & 672 x 512 & ER, M & $\approx$ 6480 & $\approx$ 720 & $\approx$ 1800\\
  $Exp_{6}$ & 672 x 512 & ER, M, C, N & $\approx$ 12710 & $\approx$ 1410 & $\approx$ 3640 \\
  \bottomrule
\end{tabular}
}
\vspace{-17pt} 
\end{table*}

\begin{table*}[htb!]
\caption{Training time, mAP and average loss error [Time/mAP/avg loss] at the end of the training phase for each experiment. }\label{tab:train_results} \centering
\begin{tabular}{lccc}
  \toprule
   & 1-class & 2-classes & 4-classes \\
  \midrule
  Full-size image & 1h20 / 91\% / 15.02 & 2h30 / 91\% / 15.54 & 5h30 / 91\% / 14.54 \\
  Quadrants of image & 0h40 / 94\% / 03.60 & 1h40 / 93\% / 03.45 & 2h50 / 93\% / 03.59 \\
  \bottomrule
\end{tabular}
\vspace{-17pt} 
\end{table*}

\section{Experimental Design\label{sec:Experimental_Design}}
In addition to building an end-to-end process for fluorescence microscopy images of yeast cells, this paper aims  to  evaluate  the detection and classification performance of the state-of-the-art YOLOv4 algorithm on individual small objects.
Here, we use five-fold cross validation, where each time one fold is used to test the model and the remaining folds are used to train the model. From the training set we randomly selected 10\% of the train images to be used for validating the model during the training phase.
We used the dataset introduced in section \ref{sec:Data} to evaluate YOLOv4. 
Based on the numbers of unique images shown in Table \ref{tab:unique_organelle}, we determine to assess the capability of YOLOv4 to classify single- and multi-class objects using 6 various experiments as defined in Table \ref{tab:images_total}. In this Table, the approximate number of images in each fold in the train-, validate-and test sets for each experiment are shown.
It is to note that YOLOv4 was trained on original images sizes ($Exp_{1}$, $Exp_{2}$ and $Exp_{3}$) versus quadrant of the images ($Exp_{4}$, $Exp_{5}$ and $Exp_{6}$). 
On average, 187k, 306k and 575k individual cells has been cropped for $Exp_{1}$, $Exp_{2}$ and $Exp_{3}$, respectively, while for $Exp_{4}$, $Exp_{5}$ and $Exp_{6}$, 173k, 284k and 539k individual cells has been cropped on average. Since the cells on the border of the images are not considered, less cells are cropped for quadrant of the images compared to full-size images. 

%
%
%
%
\section{Results and Discussion\label{sec:Results_Discussion}}
\begin{figure*}[htb!]
\centering
\includegraphics[width=0.90\textwidth]{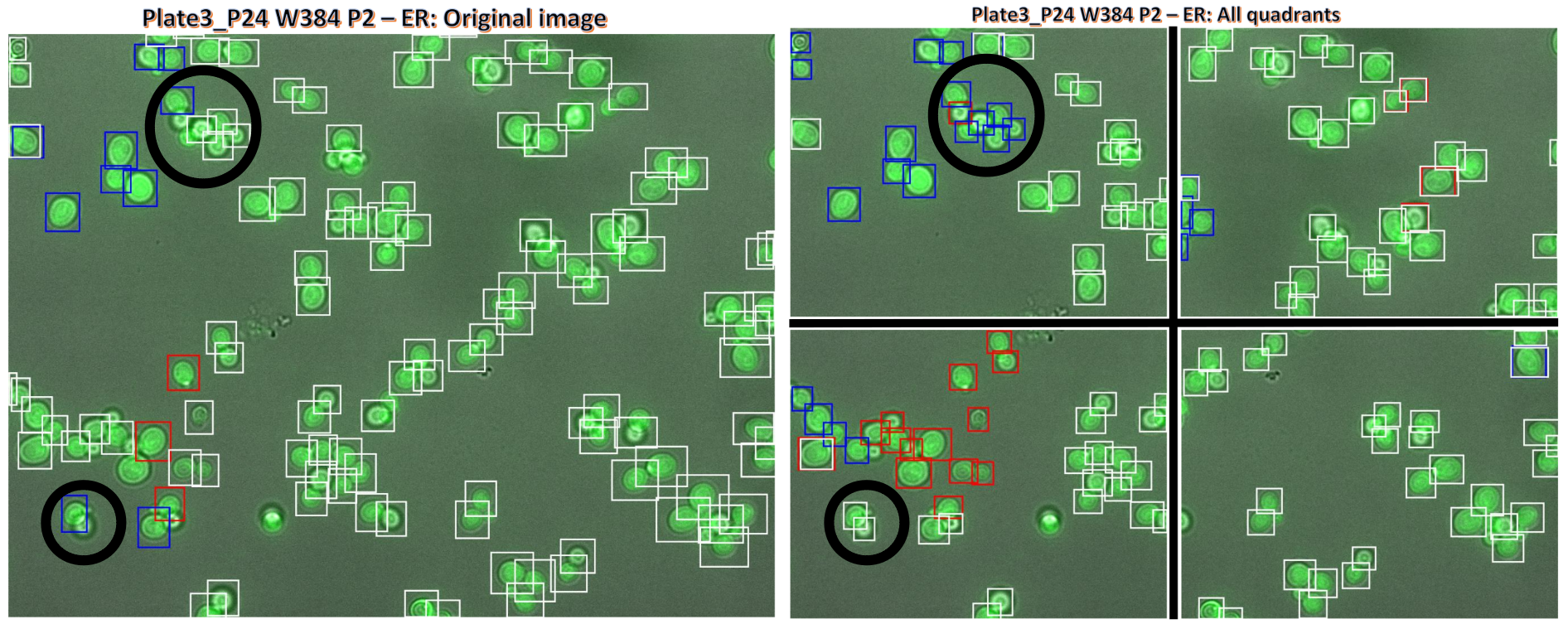}
    \caption{Detection and classification results for an image [Plate3\_P24] using both the complete, subsampled image and the native-resolution quadrants. 
    \label{FIG:BadImage-quadrants-original}}
    \vspace{-17pt} 
\end{figure*}

\begin{figure}[!htb]
\centering
\includegraphics[width=0.8\textwidth]{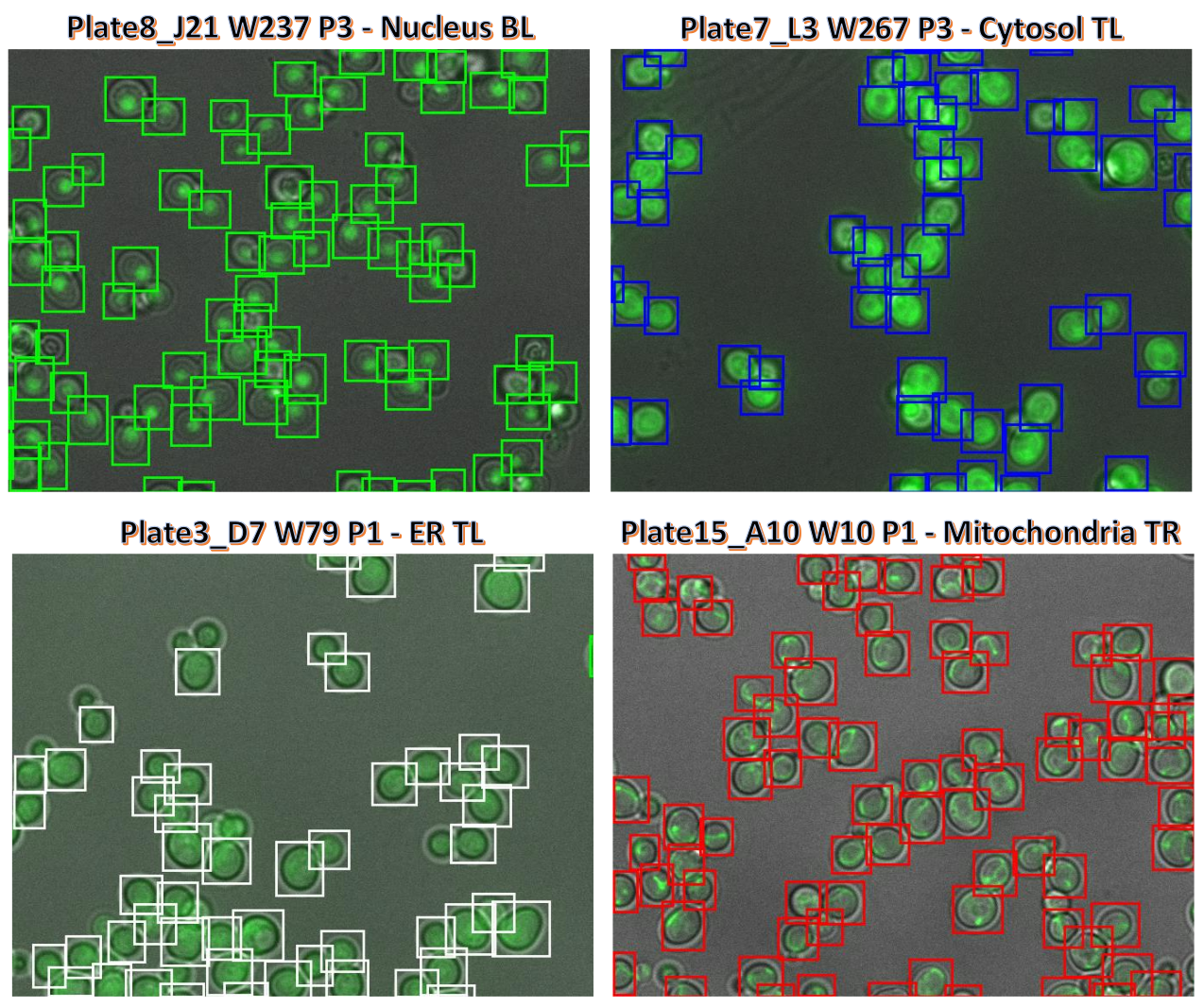}
    \caption{Randomly selected test images for 4-classes [ER, Mitochondria, Cytosol and Nucleus] classification and by using only quadrant of the images. 
    \label{FIG:4-classes-quadrant-results}}
    \vspace{-22.5pt} 
\end{figure}

This section presents and analyses the results obtained from 6 trained YOLOv4 models defined in section \ref{sec:Experimental_Design}. Each model describes one experiment and is obtained by employing the end-to-end process on the introduced dataset from section \ref{sec:Data}. 
Table \ref{tab:train_results} reports the average training time, the average mAP\footnote{Mean average precision} and the average loss error computed on the corresponding validation set for all folds. 
As it can be seen, the mAPs computed on quadrant of the images for all classification type ($Exp_{4}$,  $Exp_{5}$ and $Exp_{6}$) are higher than the mAPs computed on full size of the original images ($Exp_{1}$,  $Exp_{2}$ and $Exp_{3}$). This indicates that the detection on quadrant of the images works better than on the full-size image. 
Average loss error computed for quadrant of the images is around 3.5, which is way lower than for full-size images. 
\begin{table}
    \footnotesize
    \caption{Test results for 5-fold cross validation on four classes for full-size images (left) and quadrant images (right).}\label{tab:result_5folf_4_classes_complete_quadrant}
    \begin{tabular}{lcccc}
      \toprule
      Fold & Precision & Recall & F1 & Accuracy \\
      \midrule
    0 & 0.989 & 0.989 & 0.989 & 0.989 \\
    1 & 0.984 & 0.984 & 0.984 & 0.984 \\
    2 & 0.987 & 0.986 & 0.987 & 0.986 \\
    3 & 0.984 & 0.984 & 0.984 & 0.984 \\
    4 & 0.978 & 0.978 & 0.980 & 0.980 \\
    \textbf{AVG} & \textbf{0.985} & \textbf{0.985} & \textbf{0.985} & \textbf{0.985} \\
      \bottomrule
    \end{tabular}
    \hfill
    \begin{tabular}{lcccc}
      \toprule
      Fold & Precision & Recall & F1 & Accuracy \\
      \midrule
    0 & 0.980 & 0.980 & 0.980 & 0.980 \\
    1 & 0.974 & 0.974 & 0.974 & 0.974 \\
    2 & 0.978 & 0.978 & 0.978 & 0.978 \\
    3 & 0.973 & 0.973 & 0.973 & 0.973 \\
    4 & 0.965 & 0.965 & 0.965 & 0.965 \\
    \textbf{AVG} & \textbf{0.974} & \textbf{0.974} & \textbf{0.974} & \textbf{0.974} \\
      \bottomrule
    \end{tabular}
    \vspace{-17pt} 
\end{table}
This implies that the classification of quadrant of the images is better than on the complete images. 
Both of these observations suggest that YOLOv4 is able to detect and classify small objects best in native resolution, as opposed to a complete but subsampled image. \\
Table \ref{tab:result_5folf_4_classes_complete_quadrant} reports the average precision, recall, F1-score and accuracy computed for each test fold for $Exp_{3}$ (left) and $Exp_{6}$ (right). The last row represents the average for all folds. Using the cross validation trick, it is evident that YOLOv4 is robust since its performance on various parts of the data is similar.
Obviously, the outcomes of all these measures show that the classification of individual cells on full-size images is 1\% better than on quadrant of the images. The reason for this is because less cells are detected on original image compared to quadrants of the image. This can be seen in the black circles in Figure \ref{FIG:BadImage-quadrants-original}, where the left side of this Figure shows the detection of individual cells on the original image, and and right side shows the detection on each quadrant of the image.
In addition, the labeling for each individual cell is not obtained from human experts but from the label class of the original image. A cell where the nucleus is seen more sharply than the ER should be called nucleus, even if the plate label is ER.
\vspace{-17pt} 
\\
In Figure \ref{FIG:4-classes-quadrant-results}, the detection results of randomly selected test images from $Exp_{6}$ are shown. 
It is apparent that all images have different brightness in their background, but this is not necessarily the general case in our dataset.  The title of the images contains information about the plate number, position in the plate, cell compartment type, and cropped position from the original images. The latter can be TL, TR, BL, BR, which corresponds to top-left, top-right, bottom-left, and bottom-right, respectively. Clearly, YOLOv4 demonstrates good detection results on these images.
It is also capable of detecting parts of cells found on the border of these images. 
However, the most remarkable result to emerge from the data is that the detection of the individual cells for ER, in contrast to other classes, has much lower accuracy. As it can be seen from bottom-left of Figure \ref{FIG:4-classes-quadrant-results}, YOLOv4 fails to detect the tiny cells.
According to biologists, ER in yeast always surrounds the nucleus. Therefore, differentiating it from a nuclear signal is not so easy.
To demonstrate this, Table \ref{tab:ConfusionMatrix} represents the normalized confusion matrix of one fold of $Exp_6$. Here, we count the number of the true prediction and the negative prediction for all cell compartments. For example, for all ER images, we count the number of ER, Cytosol, Mitochondria and Nucleus predicted cell compartments, respectively. In this case, the latter three classes are considered the false prediction. Finally, we normalize all these four values, and show them in the first row of the normalized confusion matrix. From Table \ref{tab:ConfusionMatrix}, it is noticeable that the classification for classes Nucleus and Mitochondria are the best with a 99\% correct prediction, while the classification for ER is the worst with only 93\% correct individual cells prediction. The red text in Table \ref{tab:ConfusionMatrix} supports the previous assumption, since 3\% of individual ER cells are predicted as Nucleus.

\begin{table}[htb!]
\caption{Confusion matrix for one fold in $Exp_{6}$.}\label{tab:ConfusionMatrix} \centering
\resizebox{0.4\textwidth}{!}{%
\begin{tabular}{clcccc}

  & & \multicolumn{4}{c}{Predicted}\\
 %
  \multirow{5}{*}{\rotatebox[origin=c]{90}{True}} & & ER & C & M & N\\
  \cline{3-6}
& ER & \multicolumn{1}{|l}{} 0.931 & 0.022 & 0.012 &  \multicolumn{1}{l|}{\textcolor{red}{0.034}}  \\
& C & \multicolumn{1}{|l}{} 0.005 & 0.969 & 0.001 &  \multicolumn{1}{l|}{0.025}  \\
& M & \multicolumn{1}{|l}{} 0.001 & 0.001 & 0.995 &  \multicolumn{1}{l|}{0.002}  \\
& N & \multicolumn{1}{|l}{} 0.003 & 0.005 & 0.001 &  \multicolumn{1}{l|}{0.991}  \\
  \cline{3-6}
\end{tabular}
}
\vspace{-17pt} 
\end{table}

\noindent Further analysis shows that the classification results using majority vote of all quadrants to classify the whole plate are similar to using the majority to classify the full-size image. 
As previously mentioned, less cell compartments are detected when using full-size images as input for YOLOv4 ($Exp_{1}$, $Exp_{2}$ and $Exp_{3}$). 
Combining this result with the previous presented results, we deduce that the trick used to divide the images into 4 quadrants reveals better results when taking the training speed into account.
Accordingly, the outcome of all parts of the images obtained from YOLOv4 can be combined and presented as one final result. 
All results shown in this section are presented to the user at the end of the testing phase of our end-to-end process.

%
%
%
%
\section{Conclusion and Future Work\label{sec:Conclusion_Future_Work}}
We presented our developed fully automated end-to-end process that employs methods from deep learning: Mask R-CNN for segmentation and YOLOv4 for detection and classification. This end-to-end system is designed for biologists, who are interested in performing any segmentation, detection or classification tasks with only a limited knowledge in the deep learning field.
Although the application domain is optical microscopy in yeast cells, the method is also applicable to multiple-cell images in medical applications. 
Moreover, we evaluated the detection and classification performance of YOLOv4 on fluorescence microscopy images from the NOP1pr-GFP-SWAT library. We chose these images as they contain tiny cell compartments that are hard to detect. 
The results obtained from the last version of YOLO, YOLOv4, reveal its capability of detecting and classifying tiny objects. However, it has been shown that there is still a room for improvements.
We showed that in term of accuracy and speed it is recommended to use the trick of dividing the original image into 4 quadrants, which is optimally suited for the native resolution of the microscope and current GPU memory sizes.
Our approach also works for cell images with more than two channels. 
We are currently in the process of integrating this approach in a publicly available website that can also be used by external users in addition to PerICo users.
%
%
%
%
\section*{Acknowledgements}
\noindent This project has received funding from the European Union’s Horizon 2020 research and innovation programme under the Marie Skłodowska-Curie grant agreement No 812968. 
We thank Prof. Maya Schuldiner from Weizmann Institute of Science for providing us with their data as well as with her great collaboration with the authors. We also thank Tjaša Košir from University of Groningen for her supports and clear explanations.

%
%
%
\bibliographystyle{unsrt}
\bibliography{example}

%
\end{document}